\documentclass[letterpaper]{article}
\usepackage{uai2020}
\usepackage[margin=1in]{geometry}

\usepackage{times}

\usepackage[utf8]{inputenc} 

\usepackage{hyperref}       
\usepackage{url}            
\usepackage{booktabs}       
\usepackage{amsfonts}       
\usepackage{nicefrac}       
\usepackage{microtype}      

\usepackage{amsmath}
\usepackage{amsthm}
\usepackage{array}
\usepackage{graphicx}
\usepackage{subcaption}
\usepackage{xcolor}

\usepackage[vlined, boxed, commentsnumbered]{algorithm2e}
\usepackage{algorithmicx}

\theoremstyle{definition}
\newtheorem{mydef}{Definition}

\title{Bayesian Online Prediction of Change Points}

\author{%
{\bf Diego Agudelo-España} \\
MPI for Intelligent Systems \\
\And
{\bf Sebastian Gomez-Gonzalez}  \\
MPI for Intelligent Systems \\
\And
{\bf Stefan Bauer}   \\
MPI for Intelligent Systems \\
\AND
{\bf Bernhard Schölkopf}   \\
MPI for Intelligent Systems\\
\And
{\bf Jan Peters}   \\
MPI for Intelligent Systems, TU Darmstadt\\
}

\begin{document}
\maketitle
\begin{abstract}
Online detection of instantaneous changes in the generative process of a data sequence generally focuses on retrospective inference of such change points without considering their future occurrences. We extend the Bayesian Online Change Point Detection algorithm to also infer the number of time steps until the next change point (i.e., the residual time). This enables to handle observation models which depend on the total segment duration, which is useful to model data sequences with temporal scaling. The resulting inference algorithm for \textit{segment detection} can be deployed in an online fashion, and we illustrate applications to synthetic and to two medical real-world data sets.
\end{abstract}

\section{INTRODUCTION}

An underlying assumption in time series models is often that the parameters of the data generative process remain fixed across a sequence of observations. Yet, it is not uncommon to encounter phenomena where this assumption does not hold. Switching models \cite{chiappa2014explicit}, also known as change point models, account for the time-variability in a time series model by splitting a sequence of observations into non-overlapping segments, each with fixed model parameters. Change point detection (CPD) algorithms aim to identify the boundaries of these segments, given the observed time series and an underlying predictive model (UPM) from which the segments of observations are generated.

Many of the previous approaches to CPD focus on the offline setting \cite{harchaoui2007retrospective, stephens1994bayesian}, i.e., the detection algorithm treats the observations in batch mode, and they are focused on retrospective segmentation as opposed to predictive modeling. Online methods for CPD were proposed by \cite{adams2007bayesian, fearnhead2007line} from a Bayesian perspective. The core idea behind Bayesian Online Change Point Detection (BOCPD) is to keep a probability distribution over the \textit{run length} $r_t$, i.e., the elapsed time since the most recent change point (CP). When a new observation is available, the belief over the run length is updated accordingly. Importantly, as we explain in Section \ref{ssec:BOCPD}, the updated run length distribution enables one-step-ahead prediction in contrast with the offline approaches.

However, all these approaches have in common that they do not explicitly address inference of future CPs. We argue that this is of high practical importance for certain applications. Consider the scenario of medical time series, even though real-time classification or segmentation of time series is a challenging problem and already some solutions exist in many disciplines, online planning of interventions is often the ultimate goal and has received significantly less attention. For instance, in sleep research it was shown that acoustic stimulation has positive effects (for more details we refer to Section \ref{ssec:sleep_experiment}) only if the intervention is done in the up-shift of the slow wave. Thus, not only is online labeling required but likewise a prediction and uncertainty quantification of the current location within the current segment (i.e., slow wave phase). While the EEG time series used for sleep research might look rough and irregular at first, an endogenous circadian clock and regular sleep dynamics ensure that even in this case certain personalized duration patterns indeed exist.

We propose to extend the BOCPD framework along multiple directions in this work in order to predict future CPs. In particular,
\begin{itemize}
	\item We show how to extend the original BOCPD algorithm to infer when the next CP will occur.
	\item We generalize such extension to perform \textit{segment} detection. The term \textit{segment} detection is deliberately preferred to emphasize that our approach not only infers the segment starting time but explicitly accounts for its ending position as well.
	\item We point out the connections between Hidden semi-Markov Models (HSMMs) and BOCPD and show that the proposed model is equivalent to an HSMM which, however, supports online \textit{segment} inference.
	\item The presented framework leads to a new class of UPMs that not only depend on the run length and the most recent observed data but also on the total \textit{segment duration} $d_t$. This class of UPMs is suitable for modeling time series that share the same underlying functional shape but with a different time scale (e.g., ECG data while walking and running).
	\item We propose an algorithm to learn the model hyperparameters in a supervised fashion using maximum likelihood estimation (MLE). We provide implementations for both online inference and learning algorithms.
\end{itemize}

\section{BACKGROUND}
\label{sec:Background}
In the following we provide more details about the models we are building on top of (HSMMs and BOCPD) and point out the differences concerning inference and modeling assumptions.

\subsection{HIDDEN SEMI-MARKOV MODELS}
\label{ssec:HSMMs}

Hidden semi-Markov Models (HSMMs) represent a particular way to define a probability distribution over a data sequence $\mathbf{Y} = \mathbf{y}_1, \mathbf{y}_2, \dots, \mathbf{y}_T$. As with Hidden Markov Models (HMMs), the observed data is assumed to be generated from a hidden sequence of discrete random variables $\mathbf{z} = z_1, z_2, \dots, z_T$ , where $z_t \in \{1,2,\dots,K\}$, through the so-called emission process. The semi-Markov property defines how the hidden variables are generated in a sequential manner. Namely, an HSMM explicitly models the number of consecutive time steps $d_t$ that the hidden process remains in a particular state $z_t$ through the duration model $p(d_t|z_t)$ \cite{murphy2002hidden}. HSMMs are fully parameterized by the transition model $p(z_t|z_{t-1})$ represented by the matrix $\mathbf{A} = \{A_{i, j}\}$,
the probability mass function (p.m.f.) $p(z_1)$ of the initial hidden state $\boldsymbol{\pi} = \{\pi_i\}$, the duration model induced by matrix $\mathbf{D} = \{D_{i, d}\}$ and the emission parameters $\boldsymbol{\theta} = \{\theta_i\}$. Given that the $i$-th hidden state generates a segment of total duration $d$ at time step $t$, the emission likelihood for the sequence of observations $\mathbf{Y}_{t,t+d-1} = \mathbf{y}_t, \mathbf{y}_{t+1}, \dots, \mathbf{y}_{t+d-1}$ is
\begin{equation}
\label{eq:Y_given_S}
p(\mathbf{Y}_{t,t+d-1} | \boldsymbol{\theta}, z_t=i, d_t=d).
\end{equation}
Note that the observations in the sequence $\mathbf{Y}_{t,t+d-1}$ are not necessarily conditionally independent among each other given the corresponding hidden variables. Moreover, they could even potentially depend on the total segment duration $d$. This allows to express more complex emission processes than with HMMs and to capture richer dependencies among distant observations.

\subsection{BAYESIAN ONLINE CHANGE POINT DETECTION}
\label{ssec:BOCPD}
\begin{mydef}[\textit{run length}]
	\label{def:runlength}
	Let the run length $r_t$ be a non-negative discrete variable denoting the number of time steps elapsed at time $t$ since the last CP. It holds that $r_t = 0$ at a CP.
\end{mydef}
The BOCPD \cite{adams2007bayesian} generative model assumes that the segments composing a sequence of observations $\mathbf{Y}_{1,T}$ form a product partition \cite{barry1993bayesian}. This means that every sequence block $\rho_k = [a_k,b_k]$ (i.e., segments for HSMMs) is generated from a fixed observation model $p(\mathbf{Y}_{\rho_k}| \boldsymbol{\omega}_k) = \prod_{t=a_k}^{b_k} p(\mathbf{y}_t | \boldsymbol{\omega}_k)$. The parameters $\boldsymbol{\omega}_k$ are drawn i.i.d.\ across segments from some fixed distribution parameterized by $\boldsymbol{\theta}$. Formally, the UPM represents the distribution $p(\mathbf{y}_{t+1}| r_t, \mathbf{Y}_{1,t}, \boldsymbol{\theta})$ which generates observations based on the current run length $r_t$ and observed values $\mathbf{Y}_{1,t}$. Under this model a CP occurs when the underlying generative model changes and a new segment is generated.
BOCPD relies on computing the run length posterior $p(r_t | \mathbf{Y}_{1,t})$ to find change points in an online manner. This posterior can be conveniently written in terms of the joint $p(r_t, \mathbf{Y}_{1,t})$, for which the following recursion holds:
\begin{align}
\label{eq:recursive_run_length}
p(r_t, &\mathbf{Y}_{1,t}) = \sum_{r_{t-1}} p(r_t, r_{t-1} \mathbf{Y}_{1,t})\\
&= \sum_{r_{t-1}}  p(\mathbf{y}_t| r_t, \mathbf{Y}^{r_t}) p(r_t | r_{t-1}) p(r_{t-1},  \mathbf{Y}_{1,t-1}).\nonumber
\end{align}
The predictive distribution for $\mathbf{y}_t$ in \eqref{eq:recursive_run_length} is assumed to be independent of the entire history of observations $\mathbf{Y}_{1,t}$ given the subset of observations $\mathbf{Y}^{r_t}$ since the last change point according to the run length $r_t$. The subset is empty if $r_t = 0$. The CP prior model $p(r_t | r_{t-1})$ is defined through a hazard function $H(r_{t-1}) \in [0,1]$ which parameterizes the two possible outcomes for $r_t$. Namely, $p(r_t = 0| r_{t-1}) = H(r_{t-1})$ or $p(r_t = r_{t-1} + 1| r_{t-1}) = 1 - H(r_{t-1})$.

Note that the hazard function $H(.)$ implicitly induces a distribution over the duration $d$ of the segments composing an observation sequence. In the HSMM framework this is explicitly modeled through the duration distribution $p(d_t|z_t)$ introduced in Section $\ref{ssec:HSMMs}$. For instance, a constant hazard function $H(r) = c$ induces a geometric distribution over the segment duration $d$. The latter is indeed the case for the standard HMM.

The joint probability in \eqref{eq:recursive_run_length} is not only useful for detecting change points, but can also be used to perform predictions about future observations. The predictive distribution for the next observation (shown in \eqref{eq:BOCPD_prediction}) is derived by marginalizing over the current run length posterior, which is obtained from \eqref{eq:recursive_run_length}.
\begin{align}
\label{eq:BOCPD_prediction}
p(\mathbf{y}_t | \mathbf{Y}_{1,t-1}) = \sum_{r_t} p(\mathbf{y}_t | r_t, \mathbf{Y}^{r_t}) p(r_t | \mathbf{Y}_{1,t-1}).
\end{align}

\subsection{HSMMs vs BOCPD}
Note that an HSMM (Section \ref{ssec:HSMMs}) with a single hidden state $K=1$ leads to an equivalent generative model for BOCPD. One of the key differences is that HSMMs are traditionally parameterized in terms of the total segment duration, whereas the BOCPD parameterization is based on the run length through the hazard function instead. The equivalence between the hazard function and the duration model is well-understood---refer to \cite{adams2007bayesian} for details. However, the chosen parameterization has relevant implications for the UPMs, as discussed in Section \ref{ssec:duration_UPMs}.

Another key difference between HSMMs and BOCPD is related with how inference is performed and its ultimate goal. HSMMs have been mostly used for batch processing of time series and retrospective inference of hidden states through the celebrated Forward-Backward and Viterbi algorithms \cite{rabiner1989tutorial}. On the other hand, BOCPD focuses on the filtering setting (i.e., online updating) and on events which are relevant for the most recent observation. It does so by explicitly accounting for potentially incomplete segment realizations.

\section{RELATED WORK}

The original BOCPD model \cite{adams2007bayesian} has been extended by adding more structure to the generative model, and new algorithms for learning and online inference have likewise been proposed. \cite{fearnhead2007line} and \cite{knoblauch2018spatio} handle different classes of UPMs simultaneously to describe the segments of data between CPs. Furthermore, dependencies across segments can be modeled by assuming Markov structure over either different UPMs \cite{fearnhead2009bayesian} or the generative parameters $\boldsymbol{\omega}_k$ \cite{fearnhead2011efficient}. UPMs based on Gaussian processes have been used to model temporal structure within segments \cite{saatcci2010gaussian}. More generally, non-exponential family UPMs can be used within the BOCPD framework without losing the original computational advantages by leveraging variational Bayes approximations \cite{turner2013online}. 
Recently, BOCPD was made robust to outliers using $\beta$-Divergences \cite{knoblauch2018doubly}. Regarding hyperparameter learning, the offline \cite{fearnhead2009bayesian} and online \cite{wilson2010bayesian, saatcci2010gaussian, caron2012line} settings have been considered from an unsupervised perspective. To the best of our knowledge none of these works either address the prediction of CPs or handle the subsequently introduced duration-dependent emission models (Section \ref{ssec:duration_UPMs}).

More broadly, similar change point and segmentation models have also been used under a Bayesian treatment to model concept drift \cite{bach2010bayesian} (i.e., time-varying learning tasks) and as a meta-algorithm for sequence prediction and task-segmentation for piecewise stationary observation models \cite{milan2016forget}.

\section{BAYESIAN ONLINE PREDICTION OF CHANGE POINTS}
\label{sec:online-inference}

We now extend the BOCPD inference procedure to predict future CP occurrences. First, we show how to perform such inference out-of-the-box in the original BOCPD model. Second, we consider an augmented model with a discrete number of different UPMs with Markovian dynamics (i.e., an HSMM) and extend the online inference procedure for this setting to estimate, in addition to the run length, the total segment duration, and the hidden state accounting for different UPMs. We refer to the augmented inference procedure as segment detection rather than CP detection since we jointly infer both the most recent (past) CP and the next (future) CP, which in turn define the segment enclosing boundaries. Third, we describe a new family of observation models that can be handled with the proposed approach which incorporate the total segment duration (Definition \ref{def:segment_duration}) into the emission process. We finally present a learning algorithm based on MLE and discuss computational aspects of the resulting inference procedure.

\subsection{RESIDUAL TIME PREDICTION FOR BOCPD}

\begin{mydef}[\textit{residual time}]
	\label{def:residual_time}
	Let the residual time $l_t \in \mathbb{N} \cup \{0\}$ be the number of remaining observations after time $t$ to reach the end of the segment active at $t$.
\end{mydef}

The residual time $l_t$ can be expressed in terms of the run length as the event $(r_{t+1} > 0) \land (r_{t+2} > 0) \land \dots \land (r_{t + l_t} > 0) \land (r_{t + l_t + 1} = 0)$. Given observations $\mathbf{Y}_{1,t}$, the posterior over the residual time within the BOCPD framework can be expressed in terms of the run length posterior presented in \eqref{eq:recursive_run_length} as
\begin{align}
p(l_t | \mathbf{Y}_{1,t}) = \sum_{r_t} p(l_t | r_t) p(r_t | \mathbf{Y}_{1,t}) , \label{eq:residual_time}\\
\text{with} \quad p(l_t | r_t) = H(r_t + l_t) \prod_{\gamma = r_t}^{r_t + l_t - 1} (1 - H(\gamma)), \nonumber
\end{align}
where $H(\cdot)$ denotes the aforementioned hazard function. We highlight that $p(l_t|r_t)$ allows efficient offline computation since it does not depend on the observations given the run length $r_t$. Note the assumption of independence of the observation model with respect to the residual time and the segment duration.

It can be shown that for the particular case of a constant hazard $H(r) = c$ Equation $\ref{eq:residual_time}$ becomes $p(l_t | \mathbf{Y}_{1,t}) = p(l_t) = c (1 - c)^{l_t}$; i.e., it is independent of the actual observations $\mathbf{Y}_{1,t}$. This suggests that online prediction of CPs is more meaningful for non-constant hazard settings.

\subsection{BAYESIAN ONLINE SEGMENT DETECTION}
\begin{mydef}[\textit{segment duration}]
	\label{def:segment_duration}
	Let the segment duration $d_t \in \mathbb{N}$ be the number of observations jointly emitted with the observation at time $t$.
\end{mydef}

As in the original formulation (Section \ref{ssec:BOCPD}), an observation sequence is composed of non-overlapping segments $\mathbf{Y}_{1,t} = \mathbf{Y}_{\rho_1}, \mathbf{Y}_{\rho_2}, \dots, \mathbf{Y}_{\rho_S}$. But the parameters $\boldsymbol{\omega}_k$ governing the $\rho_k$ segment are now sampled according to $
\boldsymbol{\omega}_k | z_k \sim p( {.\,} | \boldsymbol{\theta}_{z_k})$,
where the discrete random variable $z_k$ indexes a particular distribution over $\boldsymbol{\omega}_k$. We also constrain the hidden variables $z_1, z_2, \dots, z_t$ to follow a semi-Markov chain as described in \ref{ssec:HSMMs}.

Additionally, we reformulate the model in terms of two extra variables for $d_t$ (Definition \ref{def:segment_duration}) and $z_t$, in addition to the run length $r_t$ defined in Section \ref{ssec:BOCPD}.

Our goal is to compute the posterior $p(r_t, d_t, z_t | \mathbf{Y}_{1,t})$ in an online manner. It turns out that we can write a similar recursion to \eqref{eq:recursive_run_length} for the joint $\gamma_t = p(r_t, d_t, z_t, \mathbf{Y}_{1,t})$:
\begin{align}
&\gamma_t = p(r_t, d_t, z_t, \mathbf{Y}_{1,t}) \label{eq:joint_t}\\
&=\sum_{r_{t-1}} \sum_{d_{t-1}} \sum_{z_{t-1}} p(r_t, d_t, z_t, r_{t-1}, d_{t-1}, z_{t-1}, \mathbf{Y}_{1,t}) \nonumber \\
&=\sum_{r_{t-1}} \sum_{d_{t-1}} \sum_{z_{t-1}} p(r_t, d_t, z_t, \mathbf{y}_t |r_{t-1}, d_{t-1}, z_{t-1}, \mathbf{Y}_{1,t-1}) \nonumber\\
&\qquad\qquad\qquad\times p(r_{t-1}, d_{t-1}, z_{t-1}, \mathbf{Y}_{1,t-1}) \nonumber\\
&=\sum_{r_{t-1}} \sum_{d_{t-1}} \sum_{z_{t-1}} p(\mathbf{y}_t |r_t, d_t, z_t, \mathbf{Y}^{r_t}) \label{eq:UPM_term}\\
&\qquad\qquad\qquad\times p(r_t, d_t, z_t|r_{t-1}, d_{t-1}, z_{t-1}) \label{eq:hidden_dyn_term}\\
&\qquad\qquad\qquad\times \underbrace{p(r_{t-1}, d_{t-1}, z_{t-1}, \mathbf{Y}_{1,t-1})}_{\gamma_{t-1}} \label{eq:joint_t_minus_1}.
\end{align}
The UPM factor in \eqref{eq:UPM_term} accounts for the generative process of observations and only takes into account the relevant observations $\mathbf{Y}^{r_t}$ according to the run length $r_t$. The factor accounting for the underlying semi-Markov dynamics in the latent discrete process is further expanded in Equation \ref{eq:semiMarkovdyn} using a similar notation to the one used in explicit-duration modeling of Markov models \cite{chiappa2010movement}.

According to \eqref{eq:joint_t} and \eqref{eq:joint_t_minus_1} $\gamma_t$ can be written in terms of itself but with a temporal shift of one time step ($\gamma_{t-1}$). This fact enables its recursive estimation by processing one observation $\mathbf{y}_t$ at a time and updating the joint in \eqref{eq:joint_t} accordingly, i.e., online updating.
\begin{align}
\begin{split}
p(r_t, d_t, z_t|&r_{t-1}, d_{t-1}, z_{t-1}) = \overbrace{p(d_t |r_t, z_t, d_{t-1})}^{\text{duration}} \\
&\times \underbrace{p(z_t |r_t, z_{t-1})}_{\text{transition}} \times \underbrace{p(r_t |r_{t-1}, d_{t-1})}_{\text{run length}},\label{eq:semiMarkovdyn}
\end{split}
\end{align}
\begin{align*}
p(d_t |r_t, z_t, d_{t-1}) &=
\begin{cases}
p(d_t | z_t)&\text{if } r_t = 0\\
\delta(d_t = d_{t-1})&\text{otherwise},
\end{cases}\\
p(z_t |r_t, z_{t-1}) &=
\begin{cases}
p(z_t | z_{t-1})&\text{if } r_t = 0\\
\delta(z_t = z_{t-1})&\text{otherwise},
\end{cases}\\
p(r_t |r_{t-1}, d_{t-1}) &=
\begin{cases}
\delta(r_t = 0)\quad\quad\text{if } r_{t-1} = d_{t-1}-1\\
\delta(r_t = r_{t-1} + 1)\quad\text{otherwise}.
\end{cases}
\end{align*}
The $\delta(x=y)$ function is $1$ iff $x$ is equal to $y$ and $0$ otherwise. The initial state p.m.f. $p(z_1)$ is omitted for the sake of brevity but must be taken into account at the moment of defining the recursion base case $\gamma_0$.

The desired posterior distribution $p(r_t, d_t, z_t | \mathbf{Y}_{1,t})$ can be computed from \eqref{eq:joint_t} using the fact that $p(r_t, d_t, z_t | \mathbf{Y}_{1,t}) \propto p(r_t, d_t, z_t, \mathbf{Y}_{1,t})$. Given that $r_t$, $d_t$ and $z_t$ are all discrete quantities, the posterior normalization constant is computed straightforwardly as $c = p (\mathbf{Y}_{1,t}) = \sum_{r_t, d_t, z_t} p(r_t, d_t, z_t, \mathbf{Y}_{1,t})$. A similar recursion to \eqref{eq:joint_t} for $p(r_t, d_t, z_t | \mathbf{Y}_{1,t})$ can also be derived. The latter recursion behaves better in terms of numerical stability given that it is a p.m.f.; therefore, the probability mass is distributed over a discrete set, as opposed to the joint $\gamma_t$. This scaling procedure is well-known in the HMM literature \cite{rabiner1989tutorial}.

Prediction for future observations is performed similarly to the BOCPD prediction in \eqref{eq:BOCPD_prediction}, but marginalizing over the new latent variables, in addition to the run length.
\begin{align}
\begin{split}
\label{eq:BOsMSD_prediction}
p(\mathbf{y}_t | \mathbf{Y}_{1,t-1}) = \sum_{r_t} \sum_{d_t} \sum_{z_t} &p(\mathbf{y}_t |r_t, d_t, z_t, \mathbf{Y}^{r_t}) \\
&\times p(r_t, d_t, z_t | \mathbf{Y}_{1,t-1}).
\end{split}
\end{align}

\subsection{DURATION DEPENDENT UNDERLYING PREDICTIVE MODELS}
\label{ssec:duration_UPMs}

There are fundamental differences between the UPM contribution $p(\mathbf{y}_t |r_t, d_t, z_t, \mathbf{Y}^{r_t})$ derived in Equation \ref{eq:joint_t}) and $p(\mathbf{y}_t| r_t, \mathbf{Y}^{r_t})$, which is the analogous expression for BOCPD (Equation \ref{eq:recursive_run_length}). Apart from the fact that we now account for a discrete number of observation models indexed by $z_t$, we also handle UPMs which not only depend on the run length $r_t$ but could also potentially depend on the total segment duration $d_t$. A dependence on $d_t$ is present, for instance, in temporal settings where the emission process is invariant to different time scales. We give a particular example of such setting in Section \ref{ssec:synthetic_experiment}. Notice that the UPM contribution of BOCPD is a special case of ours; i.e., it implicitly assumes a single hidden state and an emission model independent of the total segment duration $d_t$ given the run length $r_t$.

Whether or not to make the UPM duration dependent is a design choice that must ultimately be informed by the data of the task at hand. We propose to use the likelihood as decision criteria; the duration is incorporated into the UPM if it leads to a higher marginal likelihood or improves the likelihood of a predictive task of interest (e.g., residual time prediction).
\subsection{HYPERPARAMETER LEARNING}
\label{sec:learning}

The described algorithm for change point online inference relies on the transition matrix $\textbf{A}$, the duration matrix $\textbf{D}$, observation model hyperparameters $\boldsymbol{\theta} = \{\boldsymbol{\theta}_z\}_{z=1}^K$ and initial state p.m.f. $\boldsymbol{\pi}$ (Section \ref{ssec:HSMMs}). Domain knowledge might be available to choose these hyperparameters, but it is usually limited for most real-world applications. We adopt a data-driven approach by using labeled observation sequences where the change point locations have been marked. We define an equivalent joint encoding of the variables $r_t, d_t$ and $z_t$ in terms of a set of binary random variables $\mathbf{S} = \{s_{i, t, d}\}$, where $s_{i, t, d} = 1$ if and only if the hidden state $i$ generates the segment of observations $\mathbf{Y}_{t-d+1,t}$ (i.e., the segment of length $d$ ending at time step $t$). The complete model likelihood can then be written as $\log p(\mathbf{Y}, \mathbf{S}) = \log p(\mathbf{Y}| \mathbf{S}) + \log p(\mathbf{S})$, where
\begin{align}
\log &p(\mathbf{S}) = \sum_{i,j = 1}^{K} \sum_{d,d' = 1}^{D} \sum_{t = d'+1}^T  (s_{i,t - d', d})(s_{j,t, d'}) \log A_{i, j} + \nonumber \\
&\sum_{i = 1}^K \sum_{d = 1}^D \sum_{t = d}^T  s_{i,t,d} \log D_{i,d} + \sum_{i = 1}^{K} \sum_{d=1}^{D} s_{i,d,d} \log \pi_i, \label{eq:segmentation_likelihood} \\
\log &p(\mathbf{Y}| \mathbf{S}) = \sum_{i = 1}^K \sum_{d = 1}^D \sum_{t = d}^T s_{i,t,d} \log p(\mathbf{Y}_{t-d+1,t} | \boldsymbol{\theta}_i, d), \label{eq:emission_likelihood}
\end{align}
and $K, D,$ and $T$ stand for the number of hidden states, maximum duration and sequence length respectively. It can be shown from Equation \eqref{eq:segmentation_likelihood} that
\begin{align}
\pi_i^\star \propto \sum_{d = 1}^{D} s_{i,d,d}, \quad D_{i,d}^\star \propto  \sum_{t = 1}^{T}  s_{i,t,d}, \nonumber \\ A_{i,j}^\star \propto \sum_{t = 2}^T \sum_{d = 1}^{D} \sum_{d' = 1}^{D} (s_{i,t - d', d})(s_{j,t, d'}), \label{eq:estimators}
\end{align}
are maximum-likelihood estimators for $\boldsymbol{\pi}, \textbf{A}$ and $\textbf{D}$. Similarly, we choose the UPM parameters $\boldsymbol{\theta}$ in such a way that \eqref{eq:emission_likelihood} is maximized, which depends on the particular choice of UPM. Note that maximum a posteriori (MAP) estimators can also be derived by assuming suitable priors for the model parameters. We emphasize that having access to a subset of labeled sequences allows us to set the hyperparameters in a principled way.

The complete-data likelihood written in \eqref{eq:segmentation_likelihood} and \eqref{eq:emission_likelihood} could be used to estimate the parameters in an unsupervised fashion using the EM algorithm assuming the random variables $\mathbf{S}$ as unobserved. In such case, the estimators in \eqref{eq:estimators} remain unchanged after replacing $s_{i,t,d}$ and  $(s_{i,t - d', d})(s_{j,t, d'})$ for their corresponding expected values (i.e., $\mathbb{E}[s_{i,t,d}]$, $\mathbb{E}[(s_{i,t - d', d})(s_{j,t, d'})]$) with respect to its conditional distribution given the current values of the hyperparameters. These conditional expectations can be efficiently computed using forward-backward algorithms for HSMMs (e.g., $\alpha$ and $\beta$ recursions in \cite{murphy2002hidden}).

\subsection{COMPUTATIONAL COST}
\label{ssec:computational_cost}
Note that a naive implementation of Equation \eqref{eq:joint_t} would be impractical. Using Dynamic Programming over the recursive structure of \eqref{eq:joint_t}, in a similar way to HSMMs inference algorithms \cite{murphy2002hidden}, we devised an update algorithm with complexity $\mathcal{O}(K^2 + D^3 K)$ per time-step for arbitrary UPMs. However, if we deal with UPMs whose likelihood function can be calculated incrementally through a set of sufficient statistics (i.e., exponential family likelihoods), then the cost per time-step reduces to $\mathcal{O}(K^2 + D^2 K)$ because $p(\mathbf{y}_t |r_t, d_t, z_t, \mathbf{Y}^{r_t})$ can be evaluated in constant time in such setting leveraging memoization. For a detailed description of the devised algorithms we refer to the Appendix\footnote{Source code and data are available at a public repository: \url{https://github.com/DiegoAE/BOSD}}.

It turns out that if the UPM is agnostic to the total segment duration $p(\mathbf{y}_t |r_t, d_t, z_t, \mathbf{Y}^{r_t}) =p(\mathbf{y}_t |r_t, z_t, \mathbf{Y}^{r_t})$, then the cost per online update is still $\mathcal{O}(K^2 + D^2K)$. We leave as an open question whether it is possible to achieve a more efficient update to exactly infer the residual time $l_t$ for this class of UPMs. An example of this type of UPM is presented in Section \ref{ssec:sleep_experiment}.
\section{RESULTS}
\label{sec:experiments}
\subsection{SYNTHETIC EXPERIMENT}
\label{ssec:synthetic_experiment}
\begin{figure}
	\begin{subfigure}{\linewidth}
		\centering
		\includegraphics[width=\linewidth]{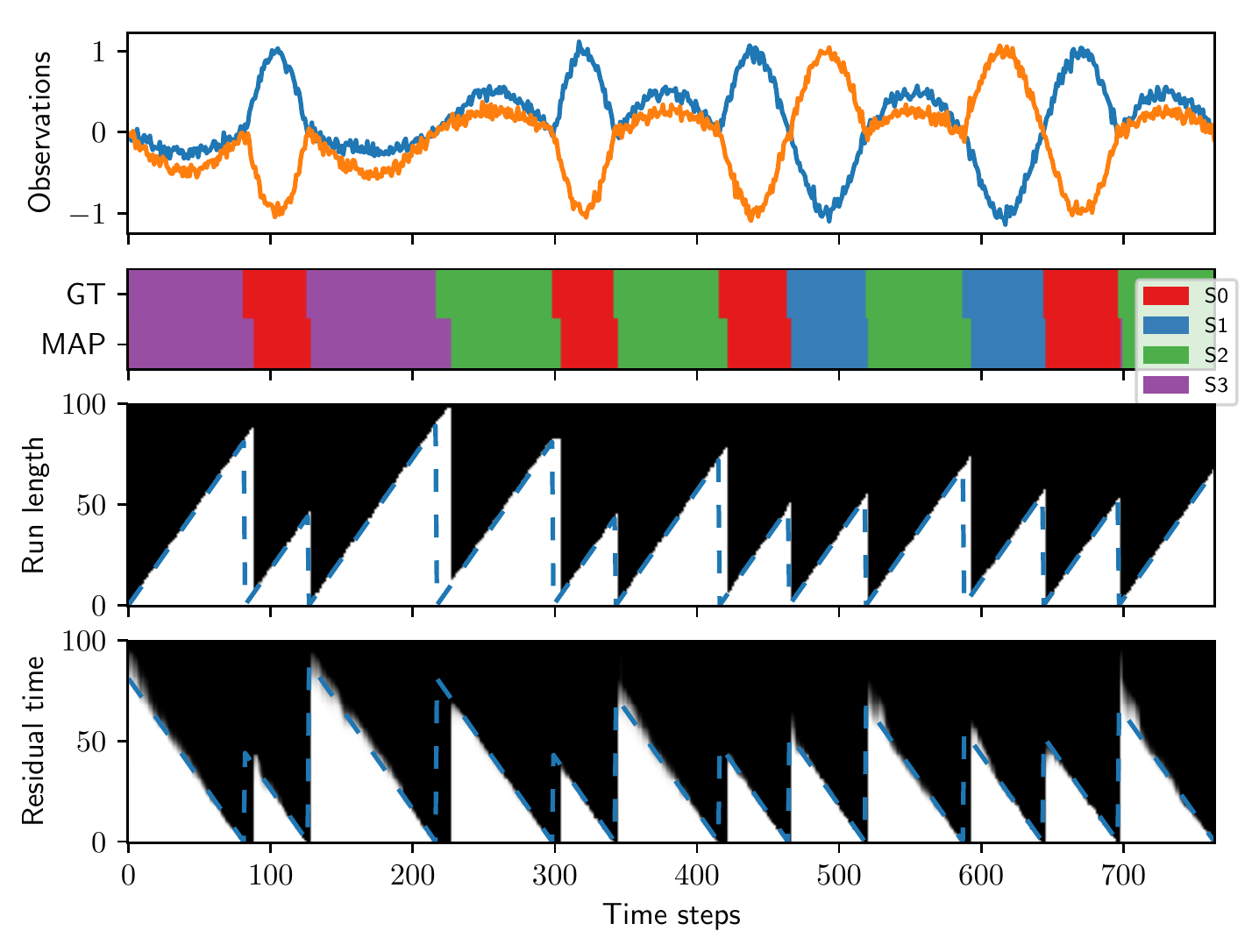}
		\caption{Synthetic experiment}
		\label{fig:synthetic_experiment}
	\end{subfigure}
	\begin{subfigure}{\linewidth}
		\centering
		\includegraphics[width=\linewidth]{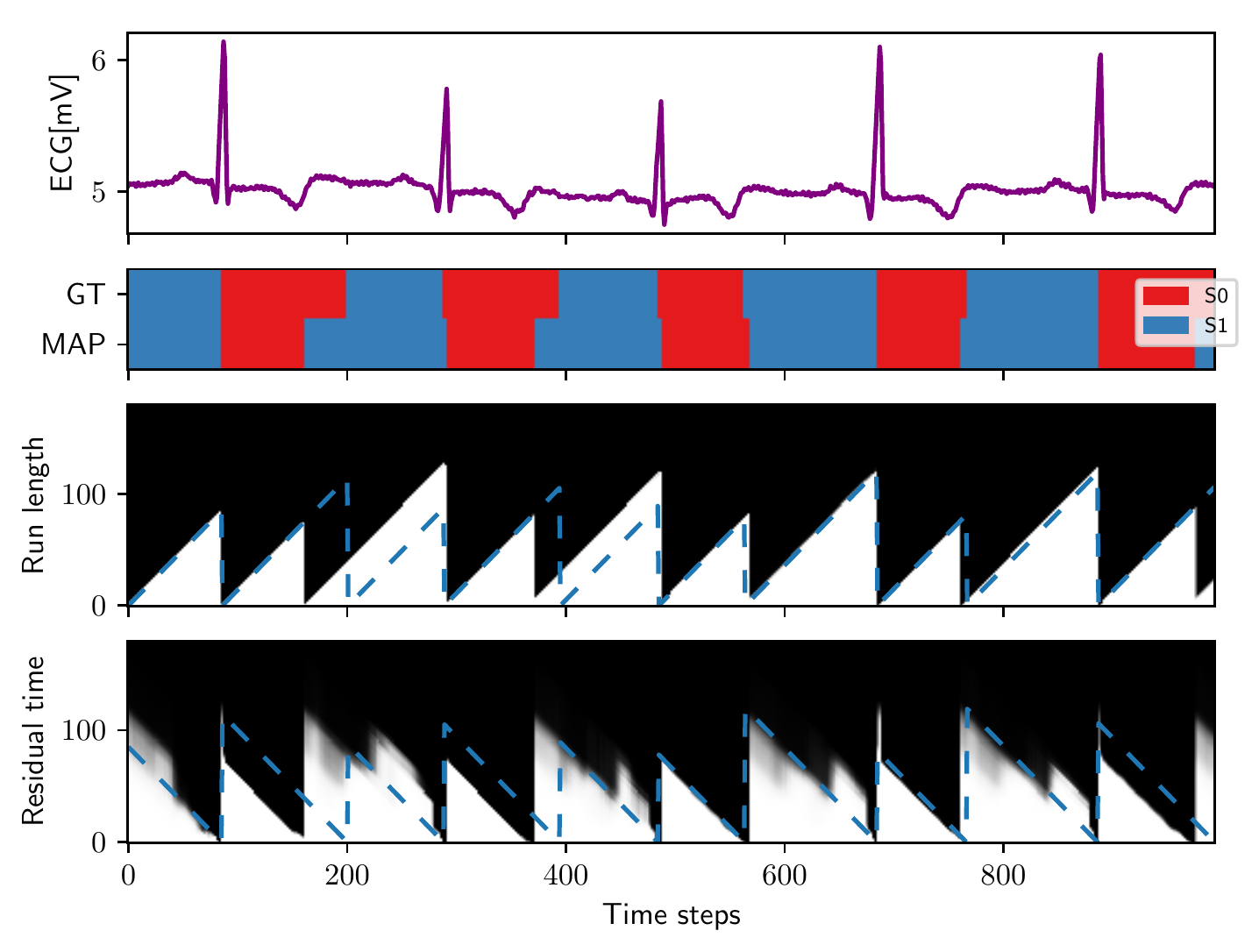}
		\caption{ECG segmentation}
		\label{fig:ecg_experiment}
	\end{subfigure}
	\caption{Inferences drawn from the proposed method in two different tasks. Both plots share the same structure where the \textbf{1st row} corresponds to the observations. The \textbf{2nd row} depicts the ground truth (GT) sequence of hidden states aligned with the sequence of states which maximizes the marginal filtering distribution (MAP)---each state is represented with a different color. \textbf{3rd and 4th rows}: the marginal run length CDF and the marginal residual time CDF are depicted as column vectors (white denotes 0 and black denotes 1). The corresponding true functions are plotted along as blue-dashed lines. Note how the true run length and remaining time functions lie in regions of high probability of the CDFs. In contrast to the experiment in Figure \ref{fig:sleep_experiment}, the shown residual time posteriors are less uncertain after seeing a few observations of a segment thanks to the duration-dependent UPM. We refer to the Appendix for larger visualizations of all plots.}
\end{figure}%
We first illustrate the proposed inference procedure using a synthetic 2D data set. The data are generated using an HSMM with 4 hidden states and with some fixed initial p.m.f. $\boldsymbol{\pi}$, transition matrix $\mathbf{A}$, and duration matrix $\mathbf{D}$. The observation model for a hidden state $k$ and for a particular segment duration of $d+1$ observations has the form $\mathbf{y}_t = (b_k \sin(t/d), c_k \sin(t/d)) + \boldsymbol{\epsilon}, \forall t \in \{0,1, \dots, d\}$ , where $\{b_k, c_k\}$ denote state-specific constants and $\boldsymbol{\epsilon} \sim \mathcal{N}(\mathbf{0}, \sigma^2\mathbf{I})$ accounts for output noise with fixed variance $\sigma^2$. Note that the sine functions input $t/d$ is always in the interval $[0,1]$ regardless of the total segment duration. This constraint encodes the property that realizations of the same hidden state with different length share the same functional shape. However, those trajectories might exhibit different time scales (e.g., linear time warping). This is an example of an observation model which jointly depends on the hidden state and the total segment duration, as described in Section \ref{ssec:duration_UPMs}. A trajectory sampled from the described model is depicted in the top row of Figure \ref{fig:synthetic_experiment}.

Note the alignment between the ground truth labels (GT) and the inferred most likely hidden state sequence (MAP) in the second row of Figure \ref{fig:synthetic_experiment}. The run length and residual time inference are also plotted alongside. Remarkably, the run length inference has almost no uncertainty as it can be seen from the sudden color change from white (0 CDF mass) to black (1 CDF mass) around the blue-dashed line depicting the true run length. This high confidence occurs as a consequence of using the actual generative models to process the synthetic observations. The run length is sometimes overestimated (e.g., around time step 300), but it gets updated to the right value after observing the first few observations of a new segment.

The residual time posterior is also fairly confident but exhibits higher uncertainty during the first observations of a new segment (e.g., time step 700). However, note that it gets more confident after seeing a few more observations. This is a consequence of the observations being dependent on the total duration of the segment they belong to. Intuitively, the very first observations in a segment can give insights about its time scale which in turn reduces the uncertainty about the remaining time until the next CP. In both cases, the inferred posteriors are consistent with the true functions of run length and residual time.
\begin{figure*}
	\centering
	\includegraphics[width=\linewidth]{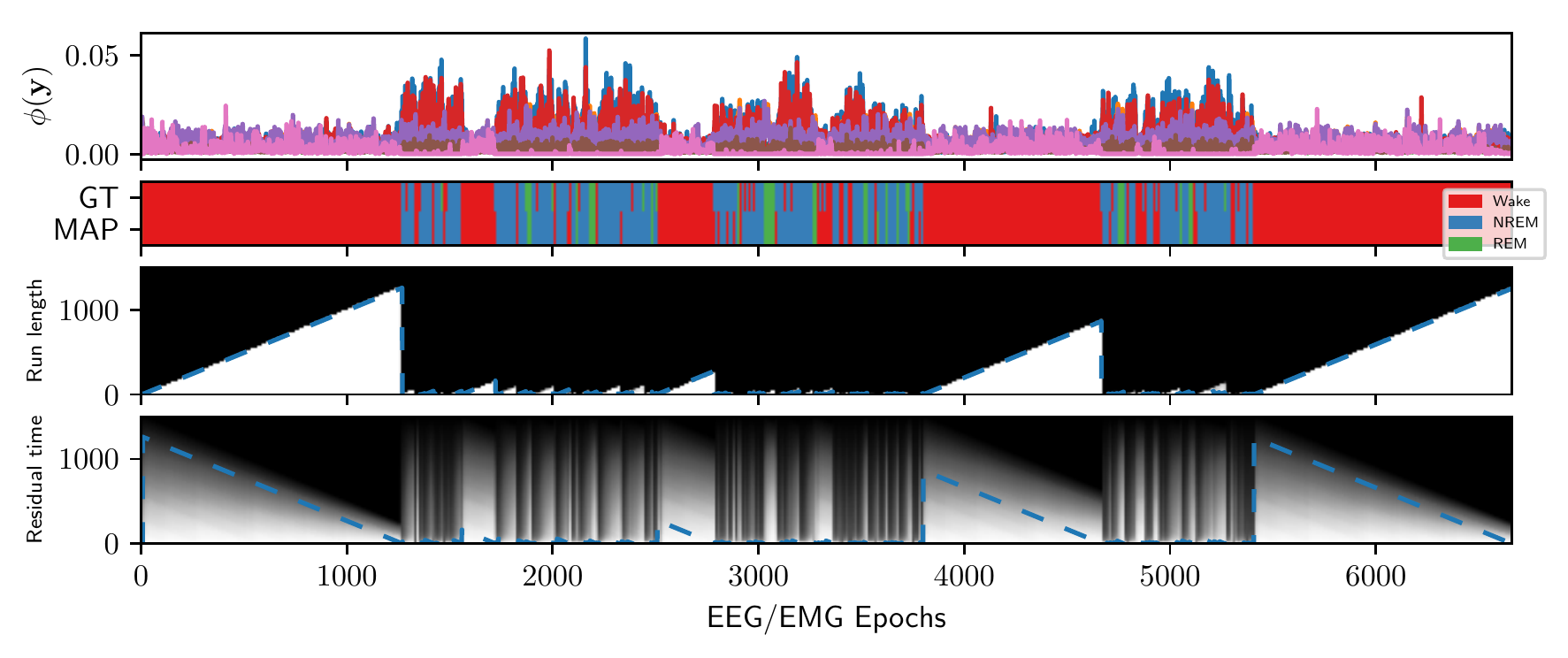}
	\caption{Mice sleep staging through Bayesian online segment detection. \textbf{First row}: shows frequency-based EEG/EMG features $\boldsymbol{\phi}(\mathbf{y}_t)$ from a validation subject. \textbf{Second row}: the sequence of sleep states provided by an expert (GT) is aligned with the sequence of states which maximizes the marginal filtering distribution $p(z_t | \mathbf{Y}_{1,t})$ (MAP). The considered sleep states are wake (red), REM (green) and NREM (blue). We recover the GT with an average F1 score of 0.91 (See Table \ref{tab:sleep_table}). In the \textbf{3rd and 4th} rows the marginal run length and marginal residual-time CDFs are plotted as columns for each epoch along ground truth functions (blue-dashed lines).}
	\label{fig:sleep_experiment}
\end{figure*}
\subsection{SLEEP CLASSIFICATION FROM EEG/EMG DATA}
\label{ssec:sleep_experiment}
Sleep staging is a fundamental but labor intensive process and numerous sleep stage classification schemes have been proposed \cite{patanaik2018end, sunagawa2013faster}. Reliable and \emph{online} sleep scoring in combination with stage dependent interventions potentially allows for acoustic stimulation \cite{ngo2013auditory}, which has been shown to have positive effects, e.g., for declarative memory. However, interventions require the phase tracking of raw EEG signals in real-time, which is in general a difficult task \cite{patanaik2018end}.

In particular, sleep staging requires identifying different sleep states from time-varying measurements; typically electroencephalography (EEG) and electromyography (EMG). In mice, sleep researchers are usually interested in distinguishing among the following states: rapid eye movement sleep (REM), non-rapid eye movement sleep (NREM) and the wake status. The standard approach to stage the EEG/EMG time series is based on frequency-band features, and it has been traditionally applied by inspection of well-trained experts.

We consider a data set with three-channel measurements (2 EEG and 1 EMG) of 3 mice recorded at a frequency of 128Hz for 24h. The data are split into 4-second epochs yielding a sequence of 21600 epochs per subject. We use one mouse as test subject and the other two and their respective labels to train the model in a supervised fashion using maximum likelihood estimation (MLE) (See Section \ref{sec:learning}).

We test our method with a maximum duration $D=1500$ and with $K=3$ hidden states. We choose an observation model which does not depend on the total segment duration for computational reasons given that it allows to handle large segment durations, as explained in Section \ref{ssec:computational_cost}. In particular, we assume the likelihood for a particular epoch $\mathbf{y}_t$ to be $p(\mathbf{y}_t | z_t, \boldsymbol{\phi}) = \mathcal{N}(\boldsymbol{\phi}(\mathbf{y}_t)| \boldsymbol{\mu}^{z_t}, \boldsymbol{\Sigma}^{z_t})$, where $\boldsymbol{\phi}(.)$ denotes a frequency-band feature mapping and $z_t$ denotes the corresponding hidden state. As features we use the amplitudes present in different frequency bands similarly to \cite{sunagawa2013faster}.

\begin{table}[htp!]
	\caption{Mice sleep state classification evaluation for the proposed method (BOSD) and FASTER.}
	\label{tab:sleep_table}
	\vskip -0.15in
	\begin{center}
		\begin{small}
			\begin{sc}
				\scalebox{0.75}{
					\begin{tabular}{l|ccc|ccc}
						\toprule
						&  \multicolumn{3}{c|}{BOSD} &  \multicolumn{3}{c}{FASTER} \\
						State & Precision & Recall & F1 & Precision & Recall & F1 \\
						\midrule
						Wake & 0.94 & 0.93 & 0.93 & 1.00 & 0.83 & 0.91\\
						REM  & 0.90 & 0.91 & 0.91 & 0.76 & 0.98 & 0.86\\
						NREM & 0.81 & 0.87 & 0.84 & 0.57 & 0.37 & 0.45\\
						\midrule
						AVG & 0.91 & 0.91 & 0.91 & 0.87 & 0.86 & 0.85\\
						\bottomrule
				\end{tabular}}
			\end{sc}
		\end{small}
	\end{center}
	\vskip -0.1in
\end{table}

In Table~\ref{tab:sleep_table}, we compare the sleep state classification performance with respect to FASTER \cite{sunagawa2013faster}, which is a fully-automated sleep staging methods designed for mice recordings. Notice that the occurrence of different sleep states is not balanced. For instance, the test sequence shown in Figure \ref{fig:sleep_experiment} has 11240 epochs labeled as wake, 8653 as NREM and only 1707 as REM. According to our experiments, FASTER does not perform well identifying the state NREM (F1 = 0.45), whereas our method gets a much higher score (F1 = 0.84) for the same class. Note that our algorithm performed online inference while FASTER considered the significantly easier offline case.

In the last two rows of Figure \ref{fig:sleep_experiment} we show the posterior CDFs over the run length and the residual time derived from the central object of our method $p(r_t, l_t, z_t | \mathbf{Y}_{1,t})$. In both cases, the probability mass is represented by a gray scale ranging from white (0 mass) to black (1 mass). The obtained run length inference is clearly much more confident than the corresponding residual time inference given the sharp pattern of its probabilities as opposed to the shadowed pattern obtained in the last row of Figure \ref{fig:sleep_experiment}. This is mostly due to the fact that the residual time prediction is inherently a predictive task, whereas the run length estimation accounts for an event that has already happened and from which there must be more evidence about. The run length estimation tends to be particularly accurate for long segments as shown by the inference performed over the consecutive runs of the red color (wake state).

Regarding the residual time prediction, we highlight that the confidence fades out in a way that is consistent with the ground truth. Consider the residual time posterior for the three longest segments of the wake state and notice how the range of plausible values is much more constrained at the end of such segments. In fact, the true remaining time function is within 2 standard deviations of the expected residual time for all the considered epochs (not shown).

In contrast with the synthetic experiment, the observations themselves can not provide much insight about the residual time in this case because in this experiment the emission process does not depend on the total segment duration. This is why we do not get a more peaked posterior over the remaining time after seeing the first few observations of a new segment.
\subsection{ECG ANALYSIS}
The measurement of the duration cycle of ECG (Electrocardiogram) waves is a useful indicator to determine abnormalities (e.g., cardiac arrhythmia) in the heart function. Moreover, within any complete ECG cycle, there are events (e.g., R peak) which encode different stages of the heart cycle, and their locations are useful markers for diagnosing heart rhythm problems, myocardial damage and even in the drug development process \cite{hughes2004markov}. We use one of the labeled ECG recordings (sel100) of the QT database \cite{laguna1997database} to showcase joint inference over the heart cycle stages (Diastole and Systole), the elapsed time on them, and the remaining time until the next stage transition in Figure \ref{fig:ecg_experiment}.

We leverage two properties of the ECG waves to fully exploit our inference procedure. Namely, bell-shaped duration models and consistent temporal patterns associated with the different heart stages. Furthermore, the functional shape of such patterns exhibit time scaling. This motivates the use of an UPM that takes the total duration into account, as explained in Section \ref{ssec:duration_UPMs}.

Formally, we model the ECG signal $y_t$ in terms of a vector of $N$ basis functions $\boldsymbol{\phi}(x)$ (parameterized as a neural network) and their corresponding weights $\boldsymbol{\omega}$, with i.i.d. Gaussian noise added. In addition, a conjugate prior is assumed over the weights $\boldsymbol{\omega} \sim \mathcal{N}(\boldsymbol{\mu}^{z_t},\boldsymbol{\Sigma}^{z_t})$, whose mean and covariance depend on the particular hidden state $z_t$ associated with the segment. Given the run length $r_t$, the segment hidden state $z_t$ and the total stage duration $d_t$, the observation model takes the form: $y_t = \boldsymbol{\phi}(r_t/d_t)^\top \boldsymbol{\omega} + \epsilon, \quad \epsilon \sim \mathcal{N}(0, \sigma_{z_t}^2)$.

Note that the basis function vector is a function of the ratio $x_t = r_t / d_t$ , therefore $0 \leq x_t < 1$ holds. Intuitively, this means that the shape of the functions generated by this emission process can be temporally stretched, and they can still be well represented by the same underlying model. A similar model is known in the robotics community as Probabilistic Movement Primitives \cite{paraschos2013probabilistic}.

In the top row of Figure \ref{fig:ecg_experiment} we depict the ECG waves for the last 5 heart cycles of the subject sequence under study. We use the first part of the sequence with its corresponding annotations to learn the basis functions $\boldsymbol{\phi}(\cdot)$, the observation model parameters $\boldsymbol{\theta} = \{\boldsymbol{\mu}^{z_t},\boldsymbol{\Sigma}^{z_t}, \sigma_{z_t}\}$, the transition matrix ($\mathbf{A}$) and the duration matrix ($\mathbf{D}$) in a supervised manner (Section \ref{sec:learning}). In the second row of Figure \ref{fig:ecg_experiment}, we show the segmentation resulting from the annotations (GT) on top of the inferred one (MAP) by maximizing the marginal state posterior. Regarding classification performance, we get $\text{precision} = 0.99$ and $\text{recall} = 0.81$ ($F1=0.89$) for S0 and $\text{precision} = 0.84$ and $\text{recall} = 0.99$ for S1 ($F1=0.91$). This shows that our method could be used to stage ECG waves in addition to detecting and predicting their change points. In the third and fourth rows, we illustrate the resulting inference over the run length and residual time after observing each ECG measurement $y_t$. The inference is consistent with the ground truth values (blue-dashed line) except for some segments located around the time step 100 and time step 300.

We emphasize that the used UPM depends on the total segment duration as opposed to the one used for the EEG in Section \ref{ssec:sleep_experiment}. The main consequence of this is that, in the former case, the observations directly influence the residual time estimate, while in the latter case this happens only indirectly via the run length posterior. The residual time inference of the first S1 segment in Figure \ref{fig:ecg_experiment} serves as an example to show how, after observing the initial part of a segment, the uncertainty reduces abruptly around the true values. Comparing the last rows of Figures \ref{fig:sleep_experiment} and \ref{fig:ecg_experiment}, it is clear that in the duration agnostic case the estimates are more conservative, i.e., they have higher uncertainty, as opposed to the posteriors obtained in the duration dependent case where the inferred posteriors reduce their uncertainty early on in the segments.
\section{CONCLUSIONS}
\label{sec:conclusions}
We extend the Bayesian Online Change Point Detection algorithm to infer when the next change point will occur in a stream of observations. We further generalize the inference procedure to perform Bayesian Online Segment Detection (i.e., joint inference of enclosing CPs) which leads to a new family of observation models that were not supported so far: UPMs that depend on the total segment duration in addition to the run length. We illustrate the method using different observation models, with synthetic and real world medical data sets. We emphasize that the modularity of the proposed method enables its composition with existing extensions of BOCPD and more flexible UPMs, which we leave for future work together with strategies to speed up inference.

\subsubsection*{Acknowledgements}

The authors thank Djordje Miladinovic for providing one of the datasets and members of the Empirical Inference department at the MPI for Intelligent Systems for proofreading and providing useful feedback.

\bibliographystyle{abbrv}
\bibliography{bosd.bib}

\onecolumn
\section*{Bayesian Online Prediction of Change Points (Supplementary Material)}
We provide detailed descriptions of the algorithms used within the proposed framework together with improved visualizations of the plots for the sake of clarity. Note that the presented plots can be reproduced using Python and following the instructions given in the software repository\footnote{Source code and data are available at a public repository. Download link: \url{https://github.com/DiegoAE/BOSD}} (README file).
\begin{figure*}[b]
	\centering
	\includegraphics[width=\linewidth]{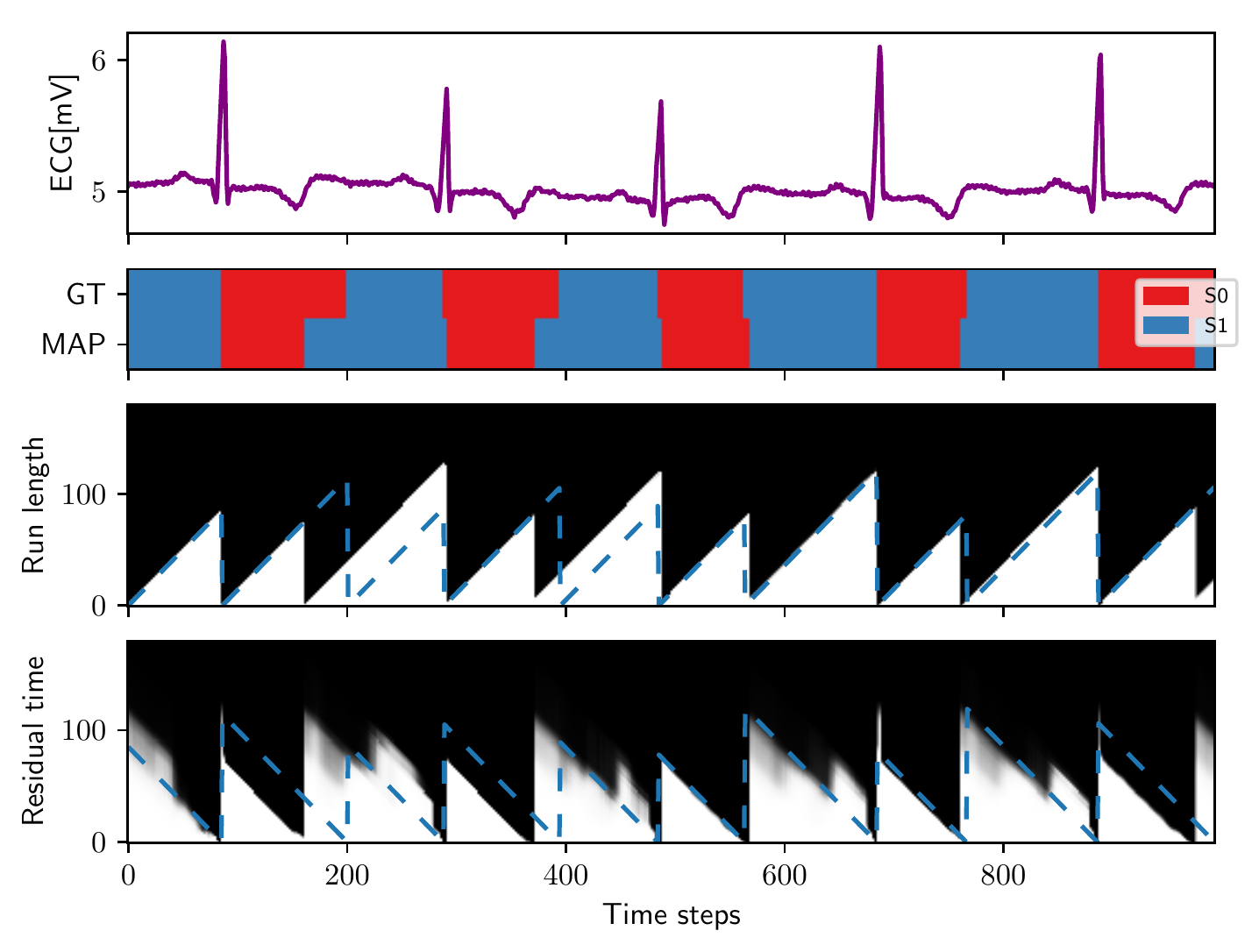}
	\caption{ECG segmentation}
\end{figure*}

\begin{figure*}
	\centering
	\includegraphics[width=\linewidth]{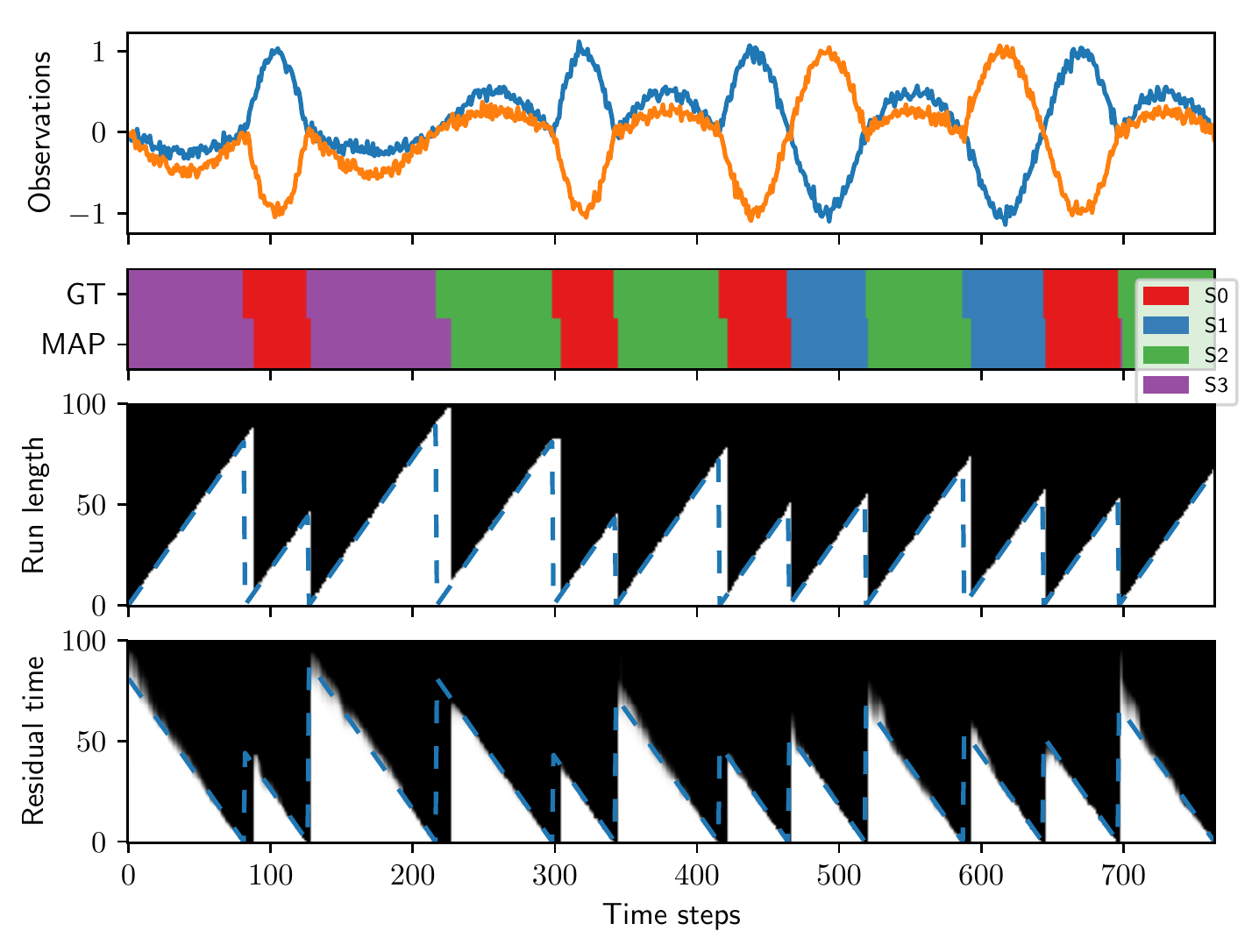}
	\caption{Synthetic experiment}
\end{figure*}

\begin{figure*}
	\centering
	\includegraphics[height=0.6\linewidth, angle=90,origin=c]{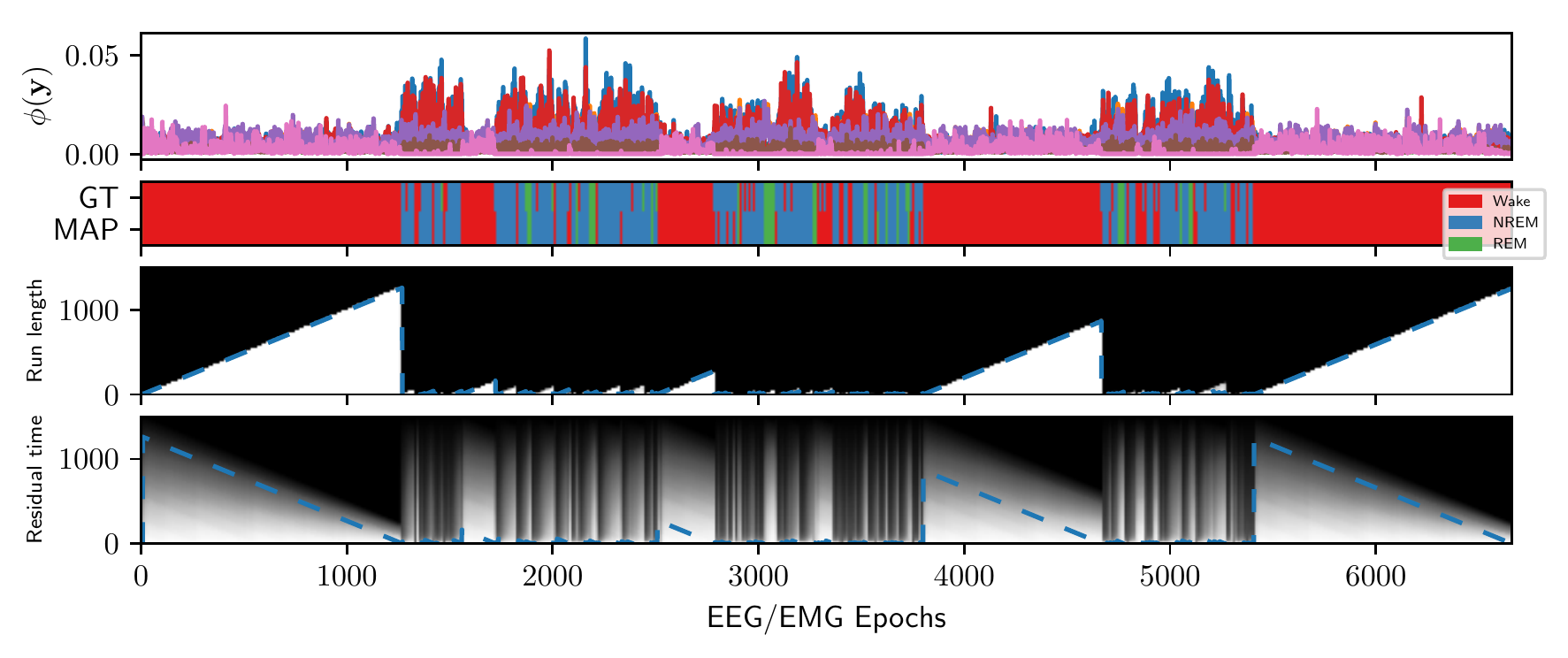}
	\caption{Mice sleep staging through Bayesian online segment detection}
\end{figure*}

\begin{algorithm*}[htbp]
	\caption{Bayesian Online Change Point Detection (BOCPD). This algorithm has complexity $\mathcal{O}(D)$ per online update. $D$ denotes the maximum total duration. Note that it is parameterized in terms of the maximum run length $R$ instead of $D$ ($R + 1 = D$).}
	\label{alg:bocpd}
	\SetKwInOut{Parameters}{Parameters}
	\Parameters{Hazard function $H(\cdot)$.\\
		Observation model prior hyperparameters $\Theta_{prior}$.}
	\For{$\text{each new observation } \mathbf{y}_t$}{
		\For{$r_t = 0, \dots, R$}{
			$\pi_t(r_t) = p(\mathbf{y}_t | \Theta_{t-1}^{r_t})$ \Comment{Compute $p(\mathbf{y}_t | {r_t}, \mathbf{Y}^{r_t})$ using sufficient statistics}\\
		}
		\For{$r_t = 1, \dots, R$}{
			$\alpha_t(r_t) = (1 - H(r_t - 1)) * \pi_t(r_t) * \gamma_{t-1}(r_t - 1)$ \Comment{Compute growth probabilities}\\
		}
		$\alpha_t(0) = 0$\\
		\For{$r_t = 0, \dots, R$}{
			$\alpha_t(0) = \alpha_t(0) + H(r_t) * \pi_t(0) * \gamma_{t-1}(r_t)$ \Comment{Compute change point probabilities}\\
		}
		$\Theta_t^0 = \Theta_{prior}$\\
		\For{$r_t = 1, \dots, R$}{
			Update $\Theta_t^{r_t}$ based on $\Theta_{t-1}^{r_t-1}$ and $\mathbf{y}_t$ \Comment{Update observation model sufficient statistics}\\
		}
		$e_t = 0$\\
		\For{$r_t = 0, \dots, R$}{
			$e_t = e_t + \alpha_t(r_t)$ \Comment{$e_t = p(\mathbf{y}_t | \mathbf{Y}_{1:t-1})$ normalizes $\alpha_t(r_t)$}\\
		}
		output $\gamma_t(r_t) = \alpha_t(r_t) / e_t$ \Comment{denotes the run length posterior $\gamma_t(r_t) = p(r_t | \mathbf{Y}_{1:t})$}\\
	}
\end{algorithm*}
\begin{algorithm*}[htbp]
	\caption{Residual time prediction for BOCPD. Note that given the run length posterior $p(r_t| \mathbf{Y}_{1:t})$ for all time steps the algorithm is independent of the actual observations. The computational complexity per online update is $\mathcal{O}(D^2)$. We have presented the run length inference (Algorithm \ref{alg:bocpd}) and the residual time inference (Algorithm \ref{alg:residual_time_bocpd}) as separate algorithms for the sake of clarity; however,  we can easily combine them within the BOCPD framework to obtain a fully online algorithm (assuming $p(l_t|r_t)$ is precomputed).}
	\label{alg:residual_time_bocpd}
	\SetKwInOut{Parameters}{Parameters}
	\Parameters{Hazard function $H(\cdot)$.\\
		Run length posterior $\gamma_t(r_t)$ computed in Algorithm \ref{alg:bocpd}.}
	\For{$r_t = 0, \dots, R$}{
		$p = 1$\\
		\For{$l_t =0, \dots, R$}{
			\If{$r_t + l_t <= R$}{
				$g(l_t, r_t) = p * H(r_t + l_t)$ \Comment{Compute $p(l_t | r_t)$ offline}\\
				$p = p * (1 - H(r_t + l_t))$
			}
		}
	}
	\For{$t = 0, \dots, T$}{
		\For{$l_t =0, \dots, R$}{
			$w_t(l_t) = 0$\\
			\For{$r_t =0, \dots, R$}{
				$w_t(l_t) = w_t(l_t) + g(l_t, r_t) * \gamma_t(r_t)$ \Comment{Recall that $\gamma(r_t)=p(r_t |\mathbf{Y}_{1:t})}$\\
			}
		}
		outputs $w_t(l_t)$ \Comment{denotes the posterior $p(l_t |\mathbf{Y}_{1:t})$}
	}
\end{algorithm*}
\begin{algorithm*}[htbp]
	\caption{Bayesian Online Segment Detection. This algorithm has complexity $\mathcal{O}(K^2 + K D^2)$, where $K$ denotes the number of hidden states. We leave as an open question whether it is possible to achieve a better complexity on D under certain conditions. As in the original BOCPD formulation we take advantage of observation models whose likelihood can be computed incrementally through a set of sufficient statistics (i.e., exponential family likelihoods). For arbitrary observation models we get $\mathcal{O}(D^2)$ complexity for BOCPD and $\mathcal{O}(K^2 + K D^3)$ for BOSD.}
	\label{alg:bosd}
	\SetKwInOut{Parameters}{Parameters}
	\Parameters{Duration matrix $\mathbf{D}(\cdot, \cdot)$ for each hidden state $z_t$.\\
		Transition matrix $\mathbf{A}(\cdot, \cdot)$\\
		Observation model prior hyperparameters $\Theta_{z_t. prior}$.}
	\For{$\text{each new observation } \mathbf{y}_t$}{
		\For{$z_t = 1, \dots, K$} {
			\For{$d_t = 1, \dots, D$} {
				\For{$r_t = 0, \dots, d_t - 1$}{
					$\pi_t(r_t, d_t, z_t) = p(\mathbf{y}_t | \Theta_{t-1}^{r_t, d_t, z_t})$ \Comment{Compute $p(\mathbf{y}_t | r_t, d_t, z_t, \mathbf{Y}^{r_t})$}\\
				}
			}
		}
		\For{$z_{t} = 1, \dots, K$} {
			\For{$d_t = 1, \dots, D$} {
				\For{$r_t = 0, \dots, d_t - 1$}{
					$\alpha_t(r_t, d_t, z_t) = \pi_t(r_t, d_t, z_t) * \gamma_{t-1}(r_t - 1, d_t, z_t)$ \Comment{Growth probabilities}\\
				}
			}
		}
		\For{$z_{t-1} = 1, \dots, K$} {
			$\eta_t(z_{t-1}) = 0$\\
			\For{$d_{t-1} = 1, \dots, D$} {
				$\eta_t(z_{t-1}) = \eta_t(z_{t-1}) + \gamma_{t-1}(d_{t-1} - 1, d_{t-1}, z_{t-1})$
			}
			\For{$z_t = 1, \dots, K$}{
				$\beta_t(z_t) = \beta_t(z_t) + \mathbf{A}(z_{t-1},z_t) * \eta_t(z_{t-1})$ \Comment{Change point probabilities}\\
			}
		}

		\For{$z_{t} = 0, \dots, K$} {
			\For{$d_t = 1, \dots, D$} {
				$\alpha_t(0, d_t, z_t) = \mathbf{D}(z_d,d_t) * \pi_t(0, d_t, z_t) * \beta_t(z_t)$ \Comment{Duration likelihood at a CP}\\
			}
		}
		Update $\Theta_t^{r_t,d_t, z_t}$ based on $\Theta_{t-1}^{r_t-1, d_t, z_t}$ and $\mathbf{y}_t$ \Comment{Update sufficient statistics}\\
		$e_t = 0$\\
		\For{$z_t = 1, \dots, K$} {
			\For{$d_t = 1, \dots, D$} {
				\For{$r_t = 0, \dots, d_t - 1$}{
					$e_t = e_t + \alpha_t(r_t, d_t, z_t)$ \Comment{$e_t = p(\mathbf{y}_t | \mathbf{Y}_{1:t-1})$ normalizes $\alpha_t(r_t, d_t, z_t)$}\\
				}
			}
		}
		output $\gamma_{t}(r_t, d_t, z_t) = \alpha_t(r_t, d_t, z_t) / e_t$ \Comment{denotes the posterior $p(r_t, d_t, z_t| \mathbf{Y}_{1:t})$}\\
	}
\end{algorithm*}

\end{document}